\documentclass[10pt]{article}

\usepackage[margin=0.72in]{geometry}
\usepackage[T1]{fontenc}
\usepackage{lmodern}
\usepackage{microtype}
\usepackage{booktabs}
\usepackage{array}
\usepackage{tabularx}
\usepackage{amsmath}
\usepackage{enumitem}
\usepackage{xcolor}
\usepackage{hyperref}
\usepackage{xurl}
\usepackage{tikz}
\usetikzlibrary{arrows.meta,fit,positioning,shapes.geometric}
\usepackage{caption}
\usepackage{titlesec}
\usepackage{placeins}
\usepackage{float}
\usepackage{graphicx}

\hypersetup{
  colorlinks=true,
  linkcolor=black,
  citecolor=green!40!black,
  urlcolor=blue!60!black,
  pdftitle={Long-Context Fine-Tuning with Limited VRAM},
  pdfauthor={Vladimir Fedosov, Aleksandr Sazhin, Artemiy Grinenko, Frank Woernle}
}
\setlist[itemize]{leftmargin=1.25em,itemsep=1.5pt,topsep=3pt}
\titleformat{\section}{\large\bfseries}{\thesection}{0.55em}{}
\titleformat{\subsection}{\normalsize\bfseries}{\thesubsection}{0.5em}{}

\title{Long-Context Fine-Tuning with Limited VRAM}
\author{Vladimir Fedosov \quad Aleksandr Sazhin \quad Artemiy Grinenko \quad Frank Woernle\\[1mm]
BMW Group\\
\url{https://github.com/vfedosov77/HierarchicalGlobalAttention}}
\date{July 2026}

\begin{document}
\maketitle
\vspace{-1.1em}

\begin{abstract}
Parameter-efficient fine-tuning reduces model and optimizer memory, but dense attention still makes long training sequences expensive. We combine Hierarchical Global Attention (HGA) with segment-wise backpropagation and tiered KV storage. Only the active segment remains differentiable in VRAM; older KV is detached into RAM or NVMe, and HGA loads a bounded set of exact historical tokens for each query block. On Qwen3-8B with 4-bit QLoRA and PG19, dense training on a 16~GB Quadro RTX~5000 fits 2,048 tokens but fails at 4,096, whereas HGA reaches 16,384 tokens with 15.28~GB peak VRAM. Under evaluation the same adapter runs through 131,072 tokens on this card; VRAM is not constant but grows gently with the resident chunk summaries, so RAM and NVMe capacity set the practical limit beyond these lengths. At the shared 2K training length, HGA-trained and dense-trained adapters obtain 2.7405 and 2.7383 nat under the same dense-attention readout, while the stock model obtains 2.9541. At this boundary HGA training is already marginally faster (217.75 vs. 207.02 tokens/s), and the HGA-to-dense throughput ratio improves from 1K to 2K; because HGA keeps the attended historical set per token approximately constant while dense work per token grows, we expect this lead to widen as context grows. Dense attention is used for the main quality and retrieval comparisons so that they measure the learned weights and remain compatible with standard generation frameworks. HGA can also be used for retrieval and generation; an optimized production-grade serving implementation is under development.
\end{abstract}

\section{Introduction}

Long-context adaptation is often limited by attention and backpropagation state rather than by the number of trainable adapter parameters. The exact boundary depends on model size, quantization, GPU capacity, and kernels, but the failure mode is general: once model weights consume most VRAM, little remains for sequence-dependent training state. QLoRA \cite{dettmers2023qlora} moves this boundary by quantizing frozen weights and training low-rank adapters, but dense attention must still process the full history.

HGA \cite{woernle2026hga} replaces a model's standard attention without changing its pretrained Q/K/V/O projections. It first routes a query block to relevant historical chunks and groups, then computes attention over the selected \emph{exact token} K/V. Here we combine HGA with truncated backpropagation through time (TBPTT) and external KV storage.

Our main results are:
\begin{itemize}
  \item HGA trains one segment at a time. Only that segment carries gradients in VRAM; a small routed subset of older tokens is read from external memory. In our setup, this raises the trainable context from 2K to 16K.
  \item At 2K, where both methods fit, HGA and dense training have the same throughput and produce practically indistinguishable adapters under a common dense-attention readout.
  \item With a fixed routing budget, HGA's historical attention work per token is approximately independent of total context length. We therefore expect HGA to become faster than dense attention on longer sequences.
  \item The resulting adapter can use either the original dense attention or HGA at inference. We use dense inference for the main comparisons and compatibility with existing serving frameworks; production-grade routed serving is ongoing work.
\end{itemize}

\section{Method}

\subsection{Exact-token hierarchical routing}

\begin{figure}[t]
\centering
\begin{tikzpicture}[
  font=\scriptsize,
  >=Latex,
  box/.style={draw,rounded corners,align=center,minimum height=8mm,inner sep=3pt},
  arrow/.style={->,thick},
  token/.style={draw,minimum width=3.4mm,minimum height=4.2mm,inner sep=0pt},
  group/.style={draw,rounded corners,inner sep=2pt}
]
  \node[font=\small\bfseries] (histlabel) {One historical chunk};
  \foreach \idx in {0,...,7} {
    \node[draw,rounded corners,fill=blue!7,minimum width=6.2mm,minimum height=5.3mm,inner sep=0pt,
          right=-\pgflinewidth of histlabel.east,xshift={8mm+6.2mm*\idx}] (grp\idx) {8};
  }
  \node[draw,rounded corners,fit=(grp0)(grp7),inner sep=3pt,
        label=above:{\textbf{64-token chunk}},label=below:{eight 8-token groups}] (chunk) {};

  \node[box,fill=gray!8,below=15mm of histlabel,xshift=5mm] (query) {current\\query block};
  \node[box,fill=green!8,right=7mm of query,minimum width=31mm] (chunksum) {all 64-token\\chunk-key summaries\\\textbf{resident in VRAM}};
  \node[box,fill=yellow!12,right=7mm of chunksum,minimum width=30mm] (groupsum) {8-token group-key\\summary cache in VRAM\\misses from RAM/NVMe};
  \node[box,fill=orange!12,right=7mm of groupsum,minimum width=29mm] (exactkv) {selected exact\\token K/V loaded\\from RAM/NVMe};
  \node[box,fill=red!7,right=7mm of exactkv,minimum width=23mm] (attn) {exact-token\\attention};

  \draw[arrow] (query) -- node[above]{score} (chunksum);
  \draw[arrow] (chunksum) -- node[above]{top-$k$} (groupsum);
  \draw[arrow] (groupsum) -- node[above]{top-$k$} (exactkv);
  \draw[arrow] (exactkv) -- (attn);
  \draw[arrow,dashed] (chunk.south) -- ++(0,-5mm) -| (chunksum.north);
\end{tikzpicture}
\caption{Two-level HGA routing. Compact chunk summaries are always available in VRAM; group summaries use a larger VRAM cache. Summaries select regions, while the final attention uses exact token K/V only. The diagram shows 8-token groups and 64-token chunks.}
\label{fig:routing}
\end{figure}
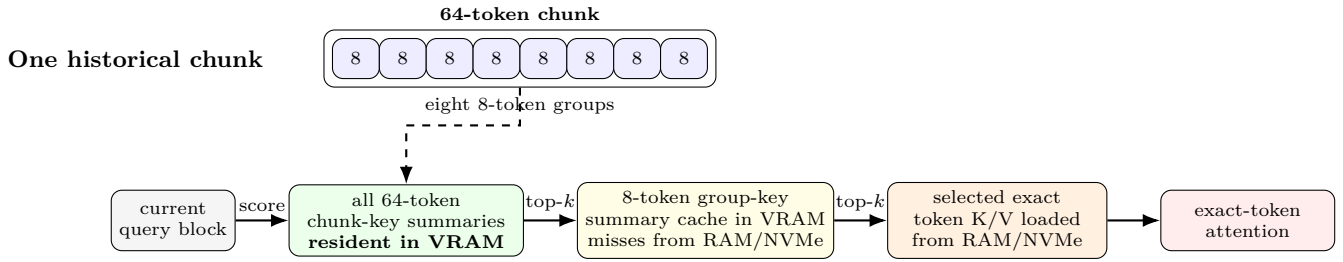

The context is divided into 64-token chunks and smaller groups; Figure~\ref{fig:routing} illustrates 8-token groups. Both sizes are configurable. HGA builds two levels of key summaries from the model's existing projected keys:
\begin{enumerate}[leftmargin=1.35em,itemsep=1pt,topsep=2pt]
  \item \textbf{Chunk selection.} A compact summary for every closed 64-token chunk remains in VRAM. The current query block scores these summaries and selects the most relevant chunks, in addition to fixed sink and local windows.
  \item \textbf{Group selection.} The selected chunks are opened at group level. Summaries of 8-token groups are kept in a larger VRAM cache; missing summaries are fetched from RAM or NVMe. The router selects the best groups inside the chosen chunks.
  \item \textbf{Exact-token attention.} Only the exact token K/V of the selected groups is loaded into the routed working set. Summary vectors are never used as attention values or output tokens.
\end{enumerate}

Routing selection runs without gradient tracking, while K/V gathering and exact-token attention remain differentiable. The wrapper keeps the pretrained projection modules by reference and adds no learned router weights, so gradients update the same Q/K/V/O and MLP projections in both dense and HGA training.

\subsection{Segment-wise backward and tiered KV storage}

A sequence of length $L$ is split into segments of length $S$. Each segment attends to its local tokens and to HGA-selected tokens from the committed history, computes the language-model loss, and immediately runs backward. At the segment boundary, its historical K/V is detached before the next segment. Later segments can read earlier content, but gradients do not cross segment boundaries. This is TBPTT \cite{williams1990tbptt}.

\begin{figure}[t]
\centering
\begin{tikzpicture}[font=\small,>=Latex]
  \node[draw,rounded corners,minimum width=5.0cm,minimum height=1.55cm,align=center,fill=blue!5] (gpu) {\textbf{GPU / VRAM}\\quantized model + adapters/optimizer\\active differentiable segment + routed exact K/V\\all chunk summaries + cached group summaries};
  \node[draw,rounded corners,minimum width=4.1cm,minimum height=1.55cm,align=center,fill=green!6,right=1.15cm of gpu] (ram) {\textbf{Host RAM}\\detached historical K/V\\nonresident group summaries\\bounded page cache for NVMe};
  \node[draw,rounded corners,minimum width=3.3cm,minimum height=1.55cm,align=center,fill=orange!8,right=1.0cm of ram] (disk) {\textbf{Filesystem / NVMe}\\full cold record\\optional spill tier};
  \draw[<->,thick] (gpu) -- node[above=1pt,fill=white,inner sep=1pt,font=\scriptsize,align=center]{exact K/V} (ram);
  \draw[<->,thick] (ram) -- node[above=1pt,fill=white,inner sep=1pt,font=\scriptsize]{pages} (disk);
\end{tikzpicture}
\caption{Tiered training state. VRAM contains the model, the active gradient segment, routed exact K/V, all chunk summaries, and a cache of group summaries. Historical capacity grows in RAM or NVMe rather than in VRAM.}
\label{fig:tiering}
\end{figure}
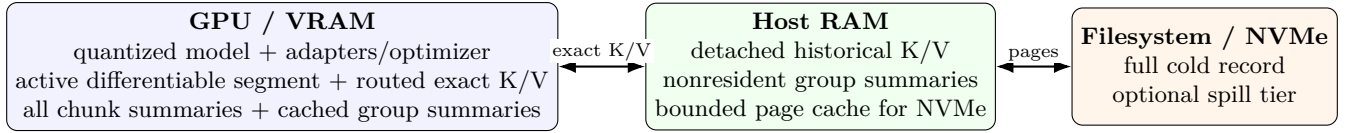

Figure~\ref{fig:tiering} summarizes the memory hierarchy. The reported experiments use RAM-backed history. The repository also contains a filesystem-backed store in which RAM is a bounded page cache over disk, suitable for NVMe.

For fixed segment length and routing budget (chunk summaries in VRAM),
\begin{equation}
M_{\mathrm{GPU}} \approx M_{\mathrm{model}} + M_{\mathrm{adapters/optimizer}} + O(S) + O(B_{\mathrm{route}}),
\end{equation}
while the external historical record grows as $O(L)$. The method therefore has no built-in 16K context constant. We tested through 16K and see no fundamental obstacle to an arbitrary context length given sufficient RAM and NVMe capacity.

\subsection{Matched dense and HGA training}

We train two QLoRA adapters with identical data order, seed, and hyperparameters. The only difference is the attention used during fine-tuning: the original dense module or HGA. Both adapters are evaluated under both attention modes. Dense readout is the primary measure of learned-weight quality; routed readout separately measures the effect of HGA selection.

\section{Experimental Setup}

The primary experiment uses Qwen3-8B \cite{yang2025qwen3}, 4-bit NF4 QLoRA, and the official PG19 train/validation splits \cite{rae2020compressive}. Quality runs use 100 optimizer steps and 32 held-out PG19 blocks.

\begin{table}[H]
\centering
\caption*{Experimental configuration.}
\small
\begin{tabular}{@{}ll@{}}
\toprule
Item & Value \\
\midrule
Model & Qwen/Qwen3-8B, 4-bit NF4 QLoRA \\
LoRA targets & Q, K, V, O, gate, up, down projections \\
Hardware & Quadro RTX 5000, 16 GB (Turing SM7.5) \\
Dataset & PG19, official train/validation splits \\
Retrieval & RULER-style passkey, multikey, multivalue \\
Adapters & qwen8b\_routed\_2k, qwen8b\_dense\_2k \\
Matched seed & 1337, identical data order \\
TBPTT segment & 2,048 tokens \\
Chunk size & 64 tokens \\
Default routing & 8 routed chunks \\
Precision / optimizer & FP16 / PagedAdamW8bit \\
Environment & \texttt{PYTORCH\_CUDA\_ALLOC\_CONF=expandable\_segments:True} \\
\bottomrule
\end{tabular}
\end{table}
\FloatBarrier

\section{Results}

\subsection{Trainable context on the same GPU}

Each context length is run in a fresh subprocess. Dense attention uses a single full-sequence forward; HGA uses 2K segments and RAM-backed history.

\begin{table}[H]
\centering
\caption{Training feasibility sweep on a 16~GB GPU. Rows through 16{,}384 keep all chunk and group summaries resident; the 32{,}768 row instead bounds the resident summary/token cache and streams the remainder from RAM, fitting the same VRAM envelope at reduced throughput.}
\label{tab:feasibility}
\small
\begin{tabular}{@{}rrrrr@{}}
\toprule
Sequence & HGA peak & HGA & Dense & Selection density \\
\midrule
2,048  & 11.76 GB & OK  & OK (11.56 GB) & 59.4\% \\
4,096  & 14.69 GB & OK  & OOM           & 32.9\% \\
8,192  & 15.26 GB & OK  & OOM           & 17.2\% \\
16,384 & 15.28 GB & OK  & OOM           & 8.7\%  \\
32,768 & 15.27 GB & OK & OOM & 4.4\%  \\
\bottomrule
\end{tabular}
\end{table}
\FloatBarrier

Dense training fails at 4K, whereas HGA completes 4K, 8K, 16K and 32K on the same card. The number of attended historical tokens grows only mildly, so the selected fraction falls from 59.4\% to 4.4\% as total context grows. The largest validated point is 32K; we do not regard it as a method limit.

\subsection{Training efficiency}

Table~\ref{tab:timing} reports a phase-by-phase comparison where both modes fit. Both measurements use the same weights; dense timing is obtained by disabling the router.

\begin{table}[H]
\centering
\caption{Per-step timing at short context (FP16). Median of ten runs after three warm-up iterations.}
\label{tab:timing}
\small
\begin{tabular}{@{}rr|rrrrrr@{}}
\toprule
Sequence & Mode & Forward (ms) & Backward (ms) & Optimizer (ms) & Total (ms) & tok/s & Peak GB \\
\midrule
1,024 & HGA   & 1,539.2 & 2,948.1 & 42.9 & 4,540.2 & 225.54 & 10.84 \\
1,024 & Dense & 1,160.0 & 2,423.6 & 48.0 & 3,627.1 & 282.32 & 10.84 \\
2,048 & HGA   & 3,224.4 & 6,077.0 & 48.8 & 9,405.3 & 217.75 & 11.76 \\
2,048 & Dense & 3,211.6 & 6,626.6 & 51.4 & 9,892.7 & 207.02 & 11.56 \\
\bottomrule
\end{tabular}
\end{table}
\FloatBarrier

At 1K, HGA is about 20\% slower (225.54 vs. 282.32 tokens/s) because routing overhead is not yet amortized and most of the short history remains selected. At 2K, HGA is already marginally faster (217.75 vs. 207.02 tokens/s), so the HGA-to-dense throughput ratio rises from 0.80 to 1.05 across this range --- the onset of the predicted crossover. For a fixed routing budget, HGA's attention work per token is approximately constant with respect to total context length, whereas dense work per token grows with the visible history; the routed selection density also keeps falling as context grows (Table~\ref{tab:feasibility}), so we expect HGA to pull further ahead at longer contexts. For large models, the transfer of a small routed working set is also expected to be small relative to the tensor operations. A direct long-context timing crossover cannot be measured on this GPU, however, because dense training already runs out of memory beyond 2K.

\subsection{Quality at equal training context}

The two adapters are trained for 100 steps at 2K with identical data order and seed. Table~\ref{tab:quality2k} cross-evaluates both adapters. The dense column is the primary comparison of learned weights; routed-minus-dense loss measures the selection cost for the same adapter.

\begin{table}[H]
\centering
\caption{Quality at 2,048-token context on 32 held-out PG19 blocks.}
\label{tab:quality2k}
\small
\begin{tabular}{@{}lrrrr@{}}
\toprule
Attention during training & HGA eval (loss / PPL) & Dense eval (loss / PPL) \\
\midrule
HGA   & 2.7433 / 15.538 & 2.7405 / 15.495 \\
Dense & 2.7426 / 15.528 & 2.7383 / 15.461 \\
Stock & --              & 2.9541 / 19.185 \\
\bottomrule
\end{tabular}
\end{table}
\FloatBarrier

The HGA-trained and dense-trained adapters differ by only 0.0022 nat under dense readout. Fine-tuning itself provides the larger effect: the best adapter reduces loss by 0.216 nat and perplexity by 19.4\% relative to the stock model. These results support HGA as a training-time replacement; they do not show that HGA training is better than dense training.

\subsection{Quality and routing cost as context grows}

Tables~\ref{tab:routedadapter} and \ref{tab:denseadapter} evaluate both adapters across context lengths. Beyond 4K, dense evaluation does not fit on this GPU, so those rows characterize HGA evaluation only. Different context lengths truncate PG19 documents differently; only comparisons between adapters at the same context length are meaningful.

\begin{table}[H]
\centering
\caption{Adapter trained with HGA, evaluated at increasing context.}
\label{tab:routedadapter}
\scriptsize
\begin{tabular}{@{}rrrrrr@{}}
\toprule
Context & HGA eval (loss / PPL) & Dense eval (loss / PPL) & Routing cost & Density & KV saving \\
\midrule
2,048  & 2.7432 / 15.537 & 2.7405 / 15.495 & +0.0027 & 59.7\% & 40.3\% \\
4,096  & 2.7180 / 15.150 & 2.7085 / 15.006 & +0.0096 & 33.1\% & 66.9\% \\
8,192  & 2.4339 / 11.404 & OOM              & --      & 17.3\% & 82.7\% \\
16,384 & 2.6443 / 14.074 & OOM              & --      & 8.8\%  & 91.2\% \\
32,768 & 2.7407 / 15.498 & OOM              & --      & 4.4\%  & 95.6\% \\
\bottomrule
\end{tabular}
\end{table}

\begin{table}[H]
\centering
\caption{Adapter trained with dense attention, evaluated under the same protocol.}
\label{tab:denseadapter}
\scriptsize
\begin{tabular}{@{}rrrrrr@{}}
\toprule
Context & HGA eval (loss / PPL) & Dense eval (loss / PPL) & Routing cost & Density & KV saving \\
\midrule
2,048  & 2.7426 / 15.528 & 2.7383 / 15.460 & +0.0044 & 59.7\% & 40.3\% \\
4,096  & 2.7189 / 15.164 & 2.7057 / 14.966 & +0.0132 & 33.1\% & 66.9\% \\
8,192  & 2.4364 / 11.432 & OOM              & --      & 17.2\% & 82.8\% \\
16,384 & 2.6488 / 14.138 & OOM              & --      & 8.8\%  & 91.2\% \\
32,768 & 2.7464 / 15.587 & OOM              & --      & 4.4\%  & 95.6\% \\
\bottomrule
\end{tabular}
\end{table}

\begin{table}[H]
\centering
\caption{Direct equal-context comparison under dense readout, where dense evaluation fits.}
\label{tab:densecomparison}
\small
\begin{tabular}{@{}rrr@{}}
\toprule
Context & Dense-trained PPL & HGA-trained PPL \\
\midrule
2,048 & 15.460 & 15.495 \\
4,096 & 14.966 & 15.006 \\
\bottomrule
\end{tabular}
\end{table}
\FloatBarrier

Table~\ref{tab:densecomparison} is the cleanest quality comparison: at both 2K and 4K, the two adapters agree within about 0.04 PPL. The HGA-only rows at longer contexts show that the adapter remains usable under increasingly sparse selection, but they are not a substitute for a dense baseline.

\subsection{Routing sparsity and cache behavior}

We vary the number of selected chunks at 4K. This experiment characterizes routed evaluation, not the dense-readout quality comparison above.

\begin{table}[H]
\centering
\caption{Routing sparsity sweep at 4,096 tokens on 32 held-out PG19 blocks for the HGA-trained adapter. Dense reference: loss 2.7085, PPL 15.006.}
\label{tab:sparsity}
\small
\begin{tabular}{@{}rrrrrr@{}}
\toprule
Top-$k$ chunks & HGA eval (loss / PPL) & Routing cost & Density & KV saving & Cache hit \\
\midrule
2  & 2.7530 / 15.690 & +0.0446 & 18.8\% & 81.2\% & 87.7\% \\
4  & 2.7334 / 15.385 & +0.0249 & 24.1\% & 75.9\% & 91.5\% \\
8  & 2.7180 / 15.150 & +0.0096 & 33.1\% & 66.9\% & 93.8\% \\
16 & 2.7126 / 15.068 & +0.0041 & 33.0\% & 67.0\% & 95.0\% \\
32 & 2.7108 / 15.042 & +0.0024 & 32.6\% & 67.4\% & 95.3\% \\
\bottomrule
\end{tabular}
\end{table}
\FloatBarrier

Quality is already close to saturation at the default setting of eight chunks. Reducing the budget to two chunks saves 81.2\% of KV interactions and raises perplexity by 0.54 relative to the eight-chunk default (0.65 relative to the densest 32-chunk setting). Cache hit rate remains 87.7--95.3\%, consistent with the summary/KV caching design in Figures~\ref{fig:routing} and \ref{fig:tiering}.

\subsection{Retrieval under dense attention}

We evaluate passkey, multikey, and multivalue retrieval with a RULER-style exact-match suite \cite{hsieh2024ruler}. Both adapters are read out with the original dense attention. This isolates whether HGA-based fine-tuning changed the learned retrieval behavior, rather than mixing it with routed-selection recall.

\begin{table}[H]
\centering
\caption{RULER-style retrieval recall (\%), dense-attention readout for both adapters.}
\label{tab:retrieval}
\small
\begin{tabular}{@{}llrr@{}}
\toprule
Attention during training & Task & 4,096 & 8,192 \\
\midrule
HGA   & Passkey    & 100.0 & 100.0 \\
HGA   & Multikey   & 100.0 & 100.0 \\
HGA   & Multivalue & 93.8  & 100.0 \\
Dense & Passkey    & 100.0 & 100.0 \\
Dense & Multikey   & 100.0 & 100.0 \\
Dense & Multivalue & 100.0 & 93.8  \\
\bottomrule
\end{tabular}
\end{table}
\FloatBarrier

Each adapter misses one of 96 planted values, in a different cell, so the result gives no evidence that HGA training harms dense-attention retrieval. HGA itself can also perform retrieval and generation, as demonstrated by the companion work and repository prototypes. We use dense readout here because it gives a clean training comparison and can immediately use existing optimized generation frameworks. A production-grade HGA serving engine, including efficient integration with systems such as vLLM or SGLang, is still under development.

\section{Related Work}

LoRA \cite{hu2022lora} and QLoRA \cite{dettmers2023qlora} reduce trainable-parameter and optimizer memory. Gradient checkpointing and FlashAttention \cite{dao2022flashattention} reduce activation materialization, but dense attention still performs all query-key interactions. LongLoRA \cite{chen2023longlora} uses shifted sparse attention during training and restores standard attention for inference. HGA uses content-based hierarchical selection over exact tokens and couples it to an external historical KV store. Classical TBPTT \cite{williams1990tbptt} supplies the bounded-gradient-history mechanism used across training segments.

\section{Limitations}
  
\subsection{Long-horizon causal leakage}
In our experiments, the current HGA training implementation remains stable for fine-tuning runs of up to approximately 100 million training tokens. However, substantially longer training reveals a causal side channel caused by sharing routing decisions across tokens within a chunk. In particular, the routing set selected for a chunk depends on scores produced by multiple query positions. An earlier token can therefore obtain indirect information about later tokens from which remote chunks were selected or rejected. Although this signal is initially weak, after approximately 100--200 million training tokens the model begins to exploit it to improve next-token prediction, resulting in measurable causal leakage. Consequently, the current implementation is suitable for fine-tuning within the tested training horizon, but not for pretraining a model from scratch. We have preliminary results indicating that a strictly causal routing variant can remove this limitation; its design and evaluation are left for future work.

\subsection{Experimental limitations} The main context-scaling experiment uses one model and GPU, and training lasts 100 optimizer steps. A matched dense-trained baseline is available only where dense training fits; quality at 8K--16K cannot be compared to dense training on the same device. TBPTT truncates gradients across segment boundaries, although later segments still attend to earlier exact K/V. The NVMe-backed tier is implemented but not benchmarked in this study. Finally, the routed inference path remains a research implementation rather than a production serving engine.

\section{Conclusion}

HGA, segment-wise backpropagation, and tiered KV storage separate the GPU working set from the full training history. In the reported Qwen3-8B/16~GB configuration, this extends the feasible training context from 2K to 16K while preserving dense-readout adapter quality and matching dense throughput at the overlap boundary. HGA training is validated through 16,384 tokens and HGA evaluation through 131,072 tokens on this card; the resident chunk summaries make VRAM grow gently as $O(L/c)$, so RAM and NVMe capacity set the practical limit beyond these lengths. Because the routed historical budget per token remains approximately constant while dense work per token grows, and HGA is already marginally faster at the 2K overlap, we expect it to become faster than dense attention as context grows. After fine-tuning, the adapter can use either standard dense attention with existing generation frameworks or HGA retrieval/generation; optimized production serving for the latter is ongoing work.

\end{document}